\documentclass[10pt,twocolumn,letterpaper]{article}

\usepackage{cvpr}

\definecolor{cvprblue}{rgb}{0.21,0.49,0.74}
\usepackage[pagebackref,breaklinks,colorlinks,allcolors=cvprblue]{hyperref}

\def\paperID{}
\def\confName{CVPR}
\def\confYear{2026}

\newcommand{\model}{EgoControl}

\title{\model{}: Controllable Egocentric Video Generation via 3D Full-Body Poses}

\author{
Enrico Pallotta\textsuperscript{*1,2} \quad
Sina Mokhtarzadeh Azar\textsuperscript{*1,2} \quad
Lars Doorenbos\textsuperscript{1,2} \quad
Serdar Ozsoy\textsuperscript{1,2} \\
Umar Iqbal\textsuperscript{3} \quad
Juergen Gall\textsuperscript{1,2} \\
\textsuperscript{1}University of Bonn \quad
\textsuperscript{2}Lamarr Institute for Machine Learning and Artificial Intelligence \quad
\textsuperscript{3}NVIDIA \\
\footnotesize\textsuperscript{*}Equal Contribution
}

\begin{document}
\twocolumn[{%
\renewcommand\twocolumn[1][]{#1}%
\maketitle
\vspace{-1.2cm}
\begin{center}
    \centering
    \captionsetup{type=figure}
    \includegraphics[width=1.\textwidth]{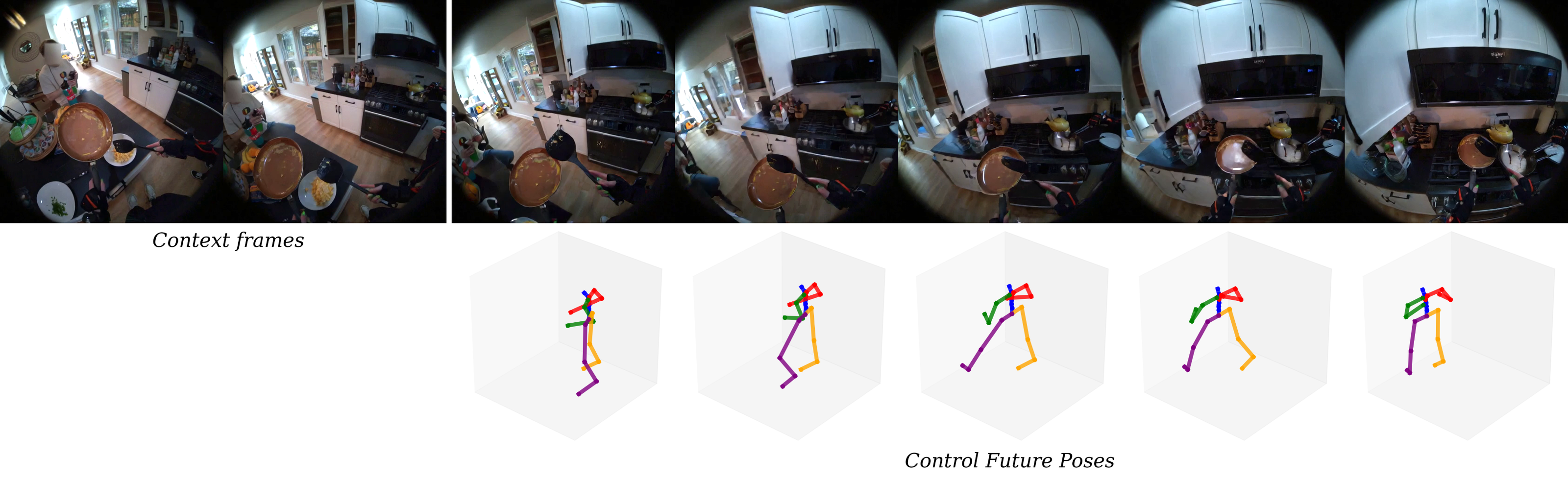}
    \vspace{-.5cm}
    \captionof{figure}{\model{} generates highly realistic egocentric videos from observed context frames and allows detailed control over the body of the camera wearer. The camera view of the generated frames follows the head pose and the visible body parts of the human pose interact with the objects in the scene in a plausible manner. Code and model available at \href{https://cvg-bonn.github.io/EgoControl/}{\texttt{cvg-bonn.github.io/EgoControl}}.}
    \label{fig:teaser}
\end{center}%
}]
\begin{abstract}
Egocentric video generation with fine-grained control through body motion is a key requirement towards embodied AI agents that can simulate, predict, and plan actions. 
In this work, we propose \model{}, a pose-controllable video diffusion model trained on egocentric data. 
We train a video prediction model to condition future frame generation on explicit 3D body pose sequences. To achieve precise motion control, we introduce a novel pose representation that captures both global camera dynamics and articulated body movements, and integrate it through a dedicated control mechanism within the diffusion process. Given a short sequence of observed frames and a sequence of target poses, \model{} generates temporally coherent and visually realistic future frames that align with the provided pose control. 
Experimental results demonstrate that \model{} produces high-quality, pose-consistent egocentric videos, paving the way toward controllable embodied video simulation and understanding. 
\end{abstract}    
\section{Introduction}
\label{sec:intro}

Egocentric (first-person) vision has recently become a central topic in computer vision~\cite{s11263-024-02095-7,THATIPELLI2025104371}. Unlike third-person imagery, egocentric video captures the sensorimotor experience of an agent, characterized by rapid camera motion, frequent occlusions from hands and objects, and a tight coupling between body pose and visual observations. This perspective is critical for problems where perception and action influence each other
, such as embodied navigation, manipulation, augmented reality, and human–robot collaboration. Consequently, the community has invested heavily in large-scale egocentric datasets and benchmarks~\cite{ma2024nymeria, grauman2022ego4d, epic-kitchen}, as well as models that can understand~\cite{lin2022egocentric, kazakos2019epic} and predict~\cite{girdhar2021anticipative} first-person behavior. These efforts have made clear that modeling the interplay between body motion and visual appearance is both highly challenging and practically important.

For an embodied agent, the ability to \emph{forecast} future egocentric frames is essential: it is a mechanism to internally simulate the visual consequences of planned actions. Predictive visual models support safer planning (by imagining risky outcomes), enable fine-grained action conditioning (by checking whether a planned motion will achieve a desired viewpoint or object configuration), and allow interactive systems (AR/VR, teleoperation, or games) to render expected views before they are executed. To be useful in these scenarios, generative models must not only be photorealistic but also controllable with fine-grained, physically plausible body-level commands that reflect how a human or a robot would move.

Existing state-of-the-art generative video models~\cite{agarwal2025cosmos, wan2025wan} are impressive at content synthesis and, more recently, at conditioning on high-level signals such as text prompts or camera trajectories~\cite{he2024cameractrl, bar2025navigation}. However, these conditioning modalities do not provide direct, explicit control over the camera wearer’s articulated body. In the egocentric setting, this is a critical gap: camera motion is produced by the wearer's head global translation and rotation, while local articulated motions (arms, hands, legs) produce the occlusions and object interactions that define first-person scenes. Without the ability to specify {full-body pose} sequences, a model cannot reliably simulate the visual consequences of specific actions (e.g., reaching, turning while walking, or performing a hand-object manipulation).

In this work, we introduce {\model{}}, the first egocentric video generative model that enables explicit control over the camera wearer’s body through 3D full-body poses. 
\model{} is a latent diffusion model trained to synthesize future egocentric frames conditioned on a short sequence of past observations and a sequence of target body poses. We design an informative pose representation that jointly captures global camera motion and articulated body dynamics, and integrate it via a dedicated control pathway within the diffusion process. This allows \model{} to accurately translate body motion into realistic egocentric visual outcomes, as shown in Fig.~\ref{fig:teaser}.

To rigorously assess pose controllability, we establish a comprehensive evaluation protocol that measures visual fidelity, global motion accuracy, and body alignment. \model{} achieves pose-consistent, high-resolution video generation. We further showcase qualitative examples demonstrating fine-grained control across diverse scenarios, including walking, turning, and hand-object interactions.

\noindent In summary, our main contributions are:
\begin{itemize}
    \item We introduce {\model{}}, a diffusion-based egocentric video generator conditioned on explicit {3D full-body pose} sequences of the camera wearer.
    \item We propose a compact and informative pose representation, along with an integrated control mechanism that jointly encodes global (camera) and articulated (body) motion to condition the diffusion process.
    \item We present a thorough quantitative and qualitative evaluation protocol designed to measure visual fidelity, camera motion accuracy, and body alignment consistency.
\end{itemize}

\section{Related Work}
\label{sec:related}
\paragraph{Egocentric vision.}
Egocentric vision has seen growing interest because first-person data provides a natural view of embodied behaviors and interactions. This surge motivated the creation of large, curated egocentric datasets and benchmarks for understanding and generation~\cite{ma2024nymeria, grauman2022ego4d, li2021egoexo, epic-kitchen}.
Work in egocentric perception covers a range of tasks, including action recognition~\cite{kazakos2019epic}, representation learning~\cite{lin2022egocentric}, dense understanding~\cite{tan2023egodistill}, and pretraining for downstream tasks~\cite{li2025egom2p}. Another important thread focuses on action anticipation~\cite{girdhar2021anticipative, zatsarynna2021multi} and early object interaction prediction from first-person streams~\cite{furnari2017next}.
Recent works address ego-pose estimation~\cite{patel2025uniegomotion} and ego$\leftrightarrow$exo generation~\cite{liu2021cross, luo2024intention, xu2025xgen}, which provide useful building blocks for egocentric generation pipelines. 

\paragraph{Video generation and prediction.}
Recent years have seen rapid progress in video generative models, driven by the rise of powerful image~\cite{rombach2022high} and video diffusion~\cite{ho2022video} approaches, as well as large-scale pretraining. 
Early efforts primarily focused on generating visually appealing videos from text prompts~\cite{ho2022imagen, singer2023makeavideo, blattmann2023stable}, while more recent approaches aim for higher temporal coherence and longer horizons.
State-of-the-art models include large-scale diffusion~\cite{yang2025cogvideox, agarwal2025cosmos} and flow-matching frameworks~\cite{polyak2024movie, wan2025wan} built upon DiT architectures~\cite{peebles2023DIT}, which currently set the benchmark for quality and realism.
\noindent In parallel, the task of {video prediction}, which involves forecasting future frames based on past observations, has been widely explored for modeling temporal dependencies and motion dynamics.
Earlier approaches employed recurrent and convolutional architectures for short- and mid-term forecasting~\cite{wang2017predrnn, gao2022simvp}, while more recent methods leverage generative diffusion models to capture uncertainty in stochastic motion~\cite{voleti2022mcvd, Pallotta_2025_CVPR}.
Motivated by similar principles as text-to-video generation, this line of work has inspired a growing family of image- and video-to-video generative models that condition on past frames, enabling more detailed and context-aware guidance during generation~\cite{hu2022make, ni2023conditional, chen2023videocrafter1, xing2024dynamicrafter}.

\paragraph{Controllable video generation.}
Recent advances in video synthesis increasingly emphasize controllability, enabling explicit conditioning on signals such as camera motion, planned trajectories, object motion, and human pose. Several methods generate videos that follow high-level camera movements~\cite{guo2024animatediff, yang2024direct} or specified camera trajectories to emulate first-person perspectives~\cite{he2024cameractrl, wang2024motionctrl, bar2025navigation, denninger2025camc2vcontextawarecontrollablevideo, zheng2024cami2v}. In autonomous-driving and navigation settings, trajectory-conditioned generators synthesize scene evolution from planned routes, demonstrating how long-range spatial cues can be integrated into diffusion pipelines~\cite{gao2024vista, hassan2025gem, russell2025gaia}. Other approaches condition on object-centric trajectories~\cite{wang2024motionctrl, shi2024motion, akkerman2025interdyn} or motion sketches~\cite{wu2024motionbooth, sudhakar2024controlling} to achieve fine-grained control of dynamic elements within a scene. Human-body pose has also emerged as a natural and effective control modality for exocentric generation: prior work guides actor motion with 2D skeletal poses using cross-attention~\cite{ma2024follow}, concatenation with the input~\cite{zhang2025mimicmotion}, or ControlNet-like mechanisms~\cite{chang2024magicpose, wang2024disco}.
Importantly, these prior methods operate under a third-person regime where the subject is largely visible and camera motion is limited. Furthermore, they only consider 2D poses, which cannot be applied to egocentric videos. In contrast, controlling generation through the \emph{egocentric} full body pose of the camera wearer presents a substantially more challenging scenario, characterized by strong viewpoint changes, frequent self-occlusions, and rich hand–object interactions. 
A first step in this direction is the concurrent work PEVA~\cite{bai2025wholebody}. It is an auto-regressive approach that generates one frame conditioned on the 3D upper-body pose per step. The approach is very limited in terms of temporal and spatial resolution.    
Our work not only enables 3D full-body pose controlled egocentric video generation at higher resolution, but it also generates videos instead of single frames, achieving temporal coherence and pose fidelity under this challenging scenario.

\section{\model{}}

\begin{figure*}
  \includegraphics[width=\linewidth]{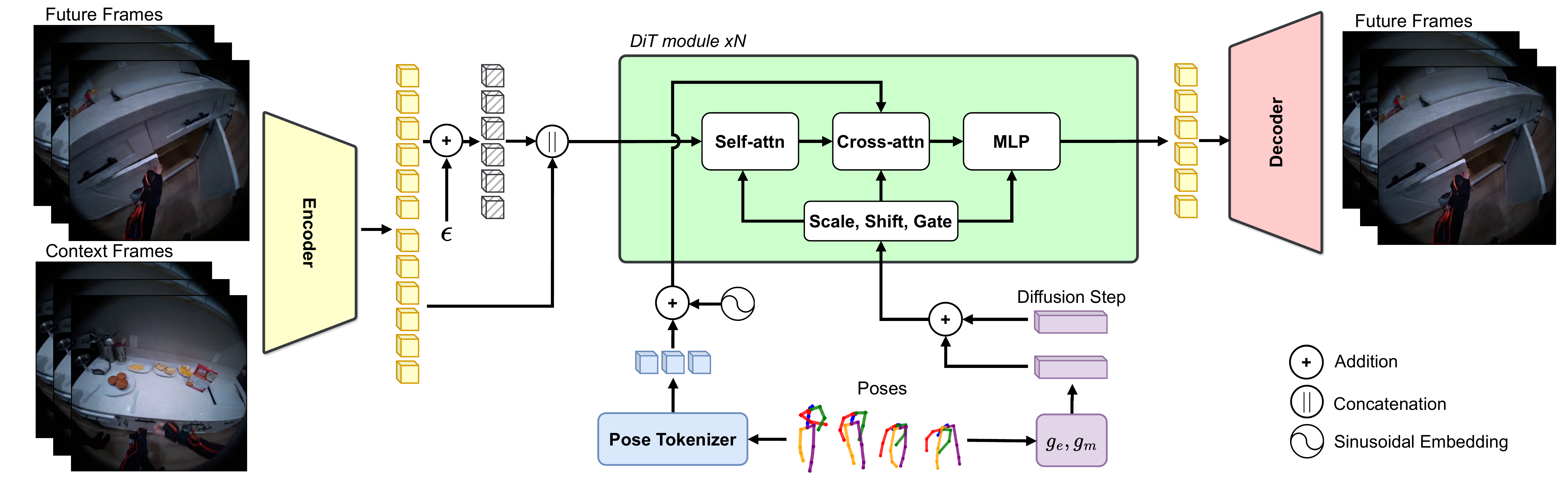}
  \caption{\model{} generates future egocentric frames conditioned on the past frames and sequence of human poses. We condition the model on the human poses in two ways. The purple branch uses the networks $g_e$ and $g_m$ to obtain modulation vectors that are combined with the ones obtained from the diffusion step. These vectors then impact the modulation by scaling, shifting, and gating the self-attention, cross-attention, and MLP blocks of DiT. The blue branch tokenizes the poses and computes the cross-attention with the encoded visual tokens. Both branches are necessary to generate realistic egocentric video frames that are well-aligned with the human pose sequence.}
  \label{fig:model}
\end{figure*}

As illustrated in Fig.~\ref{fig:teaser}, \model{} is a video diffusion model that synthesizes egocentric videos conditioned on two factors: the past visual context $\mathbf{x}{=}(x_1, x_2, \dots, x_N)$, i.e., the current observed state, and the future motion intent expressed by the human pose sequence $\mathbf{P}{=} (p_1, p_2, \dots, p_M)$. The human pose sequence not only controls the camera view, which depends on the head pose in egocentric videos, but also the body parts that are visible in the synthesized frames $\mathbf{y} {=} (y_1, y_2, \dots, y_M)$. The goal is thus to generate egocentric video frames $y_i \in \mathbf{y}$ that not only look realistic, but that are also visually coherent with their corresponding control pose $p_i \in \mathbf{P}$. 

Our method is based on a latent conditional video diffusion model \cite{agarwal2025cosmos}. Initially, a tokenizer \(\mathcal{E}\) maps frames to compact, continuous latents \(z_0=\mathcal{E}_{}(x)\). While \(q(z_t\!\mid\!z_0)\) denotes the forward perturbation process in latent space, \(p_\theta(z_{t-1}\!\mid\!z_t,c)\) denotes the learned denoiser conditioned on some context \(c\), e.g., $c=\mathbf{x}$. 
Following the EDM formulation~\cite{karras2022elucidating}, the model perturbs the clean latent using a continuous noise level \(\sigma\):
\begin{align}
    z &= z_0 + \sigma\,\varepsilon, \quad \varepsilon \sim \mathcal{N}(0, I),
\end{align}
and trains the denoiser to predict the original clean latent \(z_0\) directly:
\begin{align}
    \mathcal{L}(\theta) &= \mathbb{E}_{z_0,\varepsilon,\sigma,c}\big[\,w(\sigma)\,\|\,z_\theta(z,\sigma,c) - z_0\,\|_2^2\,\big],
\end{align}
where \(w(\sigma)\) is a weighting function controlling the relative importance of each noise level and \(z_\theta\) is the denoising network, implemented as a DiT~\cite{peebles2023DIT}.

\noindent In our work, the context \(c\) not only includes the past visual context $\mathbf{x}$ but also the future motion intent expressed by the human pose sequence $\mathbf{P}$. We discuss how we encode the human pose sequences $\mathbf{P}$ in \cref{sec:pose_representation} and describe how we effectively control the diffusion model in the challenging setting of egocentric video generation in \cref{sec:control_generation}.

\subsection{Pose Representation}
\label{sec:pose_representation}
Egocentric datasets typically provide frame-synchronized 3D body poses defined in a global reference frame. However, since \model{} is designed to operate from an embodied perspective, we require a pose representation that reflects the agent’s own motion rather than its absolute position in space. To this end, we represent each pose sequence $\mathbf{P}$ in terms of relative transformations that capture frame-to-frame motion dynamics.

Each 3D pose is given as a set of $J = 23$ transformation matrices describing the position and orientation of each joint with respect to the global frame. From an egocentric viewpoint, most perceptual changes in the video arise from the movement of the camera, which corresponds to the head motion. Therefore, given the sequence of global head poses $\mathbf{H} = (\mathbf{H}_0, \mathbf{H}_1, \dots, \mathbf{H}_M)$, where $\mathbf{H}_i \in \mathbb{R}^{4\times4}$, we compute the relative head transformation between consecutive frames as
\begin{equation}
    \Delta \mathbf{H}_i = \mathbf{H}_i \mathbf{H}^{-1}_{i-1}, \quad \forall i \in [1, M].
\end{equation}
Each $\Delta \mathbf{H}_i$ is then converted into a 6D vector $\Delta\mathbf{h}_i \in \mathbb{R}^{1\times6}$ representing translation and rotation in Euler format, yielding the sequence of head movements $\mathbf{\Delta} \mathbf{h} = (\Delta\mathbf{h}_1, \dots, \Delta\mathbf{h}_M) \in \mathbb{R}^{M\times1\times6}$.

To incorporate full-body state, we also compute the relative transformations of all joints with respect to the pelvis at each frame, denoted as $\mathbf{J} \in \mathbb{R}^{M\times21\times6}$, where $21$ is the number of joints excluding head and pelvis. Furthermore, we model the pelvis as the root joint and compute its frame-to-frame relative motion $\mathbf{\Delta} \mathbf{r} = (\Delta\mathbf{r}_1, \dots, \Delta\mathbf{r}_M) \in \mathbb{R}^{M\times1\times6}$ as we did for the head.

Finally, we concatenate these components to obtain the full pose representation that will control the generation of the $M$ future frames:
\begin{equation}
    \mathbf{P} = [\mathbf{\Delta} \mathbf{h}, \mathbf{\Delta} \mathbf{r}, \mathbf{J}] \in \mathbb{R}^{M\times23\times6}.
\end{equation}
This representation allows \model{} to capture both egocentric camera dynamics and articulated body motion in a unified form.

\subsection{Control Mechanism}
\label{sec:control_generation}

We propose to use two control mechanisms to condition the video generation on 
the full human body pose representation $\mathbf{P}$ as illustrated in \cref{fig:model}. The first approach via modulation is described in \cref{sec:pc_adaln} and the second approach via cross-attention is described in \cref{sec:pc_ca}. We show that the combination of both gives the best results.       

\subsubsection{Pose control via modulation}\label{sec:pc_adaln}

Inspired by \cite{8237429,8953676,agarwal2025cosmos}, we use a modulation and gating approach to update the features in the DiT~\cite{peebles2023DIT} blocks, where the modulation and gating parameters depend on the full-body poses.       
To this end, we first flatten the entire pose tensor $\mathbf{P} \in \mathbb{R}^{M\times23\times6}$ across time and joints, yielding a single global vector $\mathbf{P}_{\text{flat}} \in \mathbb{R}^{(M\cdot 23\cdot6)}$. A pair of two-layer MLPs $g_e$ and $g_m$ then map this vector to the embedding and modulation space:
\begin{equation}\label{eq:ep_mp}
    \mathbf{e}_P = g_e(\mathbf{P_{\text{flat}}}), \qquad 
    \mathbf{m}_P^{\beta \gamma g} = [\mathbf{m}_P^{\beta}, \mathbf{m}_P^{\gamma}, \mathbf{m}_P^{g}] = g_m(\mathbf{P_{\text{flat}}}).
\end{equation}
Given the pose embedding $\mathbf{e}_P$ and its modulation vector $\mathbf{m}_P^{\beta \gamma g}$, different shift, scale and gate parameters are predicted for the self-attention, cross-attention, and MLP components in each block:
\begin{equation}
\label{eq:adaln_params}
\begin{aligned}
[\boldsymbol{\beta}_P^{(k)}, \boldsymbol{\gamma}_P^{(k)}, \mathbf{g}_P^{(k)}]
&= \mathbf{W}^k_{m1}\mathbf{W}^k_{m2} \text{SiLU}(\mathbf{e}_P)
   + \mathbf{m}_P^{\beta \gamma g},
\end{aligned}
\end{equation}
where $k$ denotes self-attention, cross-attention, or MLP. $\mathbf{W}^k_{m1} \in \mathbb{R}^{3\cdot d \times r}$ and $\mathbf{W}^k_{m2} \in \mathbb{R}^{r \times d}$ are learnable projection matrices with $r < d$.
Each component input $\mathbf{u}$ is then normalized and modulated via adaptive layer normalization:
\begin{equation}
\text{AdaLN}^{(k)}(\mathbf{u}; \boldsymbol{\beta}_P^{(k)}, \boldsymbol{\gamma}_P^{(k)})
= \text{LN}(\mathbf{u}) \odot (1 + \boldsymbol{\gamma}_P^{(k)}) + \boldsymbol{\beta}_P^{(k)},
\end{equation}
where LN denotes layer normalization. 
The resulting outputs of the self-attention, cross-attention, and MLP components are gated and combined through residual connections:
\begin{align}
\mathbf{u} &\leftarrow \mathbf{u} + \mathbf{g}_P^{(\text{self})} \odot 
    \text{SelfAttn}\!\left(\text{AdaLN}^{(\text{self})}(\mathbf{u})\right), \\
\label{eq:corssattention}\mathbf{u} &\leftarrow \mathbf{u} + \mathbf{g}_P^{(\text{cross})} \odot 
    \text{CrossAttn}\!\left(\text{AdaLN}^{(\text{cross})}(\mathbf{u}), \mathbf{c}\right), \\
\mathbf{u} &\leftarrow \mathbf{u} + \mathbf{g}_P^{(\text{mlp})} \odot 
    \text{MLP}\!\left(\text{AdaLN}^{(\text{mlp})}(\mathbf{u})\right),
\end{align}
where $\mathbf{c}$ is the context for the cross-attention, which we will describe in \cref{sec:pc_ca}. This mechanism allows $\mathbf{P}$ to modulate normalization and residual pathways across all layers. 

As illustrated in \cref{fig:model}, this mechanism is also used to integrate the diffusion step $t$. To this end, $t$ is mapped to a continuous embedding space using a sinusoidal embedding function
\begin{equation}
    \mathbf{e}_t = \text{sinusoidal}(t) \in \mathbb{R}^{D},
\end{equation}
and we compute the modulation parameters for the diffusion step in the same way, i.e., $\mathbf{m}_t^{\beta \gamma g}=f(\mathbf{e}_t)$ where $f$ is a two-layer MLP. We then combine it with \cref{eq:ep_mp} by  
\begin{equation}
    \mathbf{e}_{P} = \mathbf{e}_t + \mathbf{e}_P, \qquad 
    \mathbf{m}_{P}^{\beta \gamma g} = \mathbf{m}_t^{\beta \gamma g} + \mathbf{m}_P^{\beta \gamma g}.
\end{equation}
This additive conditioning enables both diffusion step and body pose to jointly modulate the layer normalization and residual dynamics throughout the transformer blocks.

\subsubsection{Pose control via cross-attention}\label{sec:pc_ca}
In addition to the AdaLN-based modulation, we introduce a fine-grained control pathway by further conditioning the transformer blocks through cross-attention with body pose tokens as illustrated in \cref{fig:model}. Unlike the AdaLN branch, here we preserve the temporal structure of $\mathbf{P}$ to maintain frame-level alignment.
Each frame's pose vector $\mathbf{P}_m \in \mathbb{R}^{23\times6}$ is projected to the model’s feature space as:
\begin{equation}
    \mathbf{P}_m^{'} = \text{LayerNorm}(\text{GELU}(\mathbf{W}_p \mathbf{P}_m)) \in \mathbb{R}^{D_p},
    \end{equation}
yielding a sequence of $M$ pose tokens.
We then apply a sinusoidal positional encoding $\text{PE}(\cdot)$ to preserve the temporal information:
\begin{equation}
    \mathbf{c}_m = \text{PE}(\mathbf{P}_m^{'}).
\end{equation}
These pose tokens are then concatenated and used as context $\mathbf{c}$ for the cross-attention in \cref{eq:corssattention}.  
This cross-attention pathway provides temporally localized control signals that complement the global modulation from AdaLN, together enabling coherent and controllable egocentric video synthesis.

\begin{figure*}[t]
  \includegraphics[width=\linewidth]{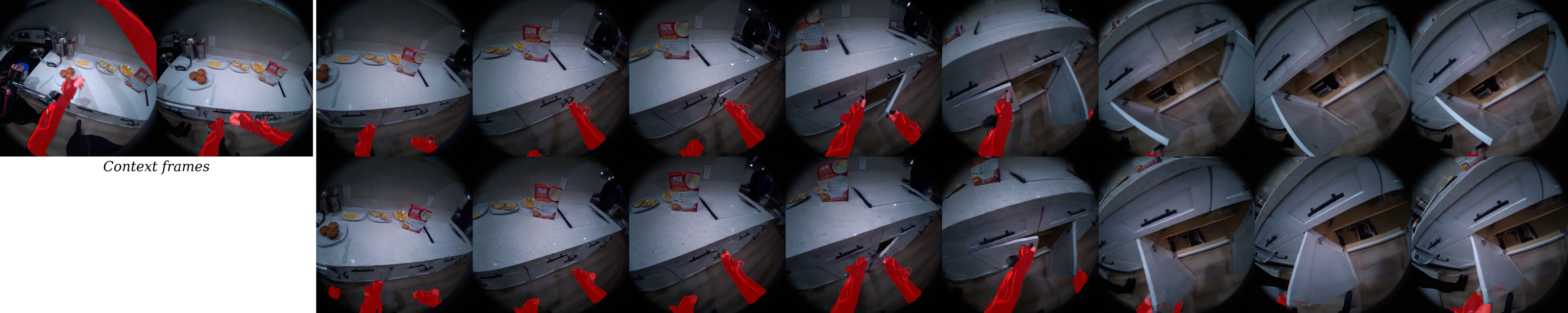}
  \caption{
  SAM2~\cite{ravi20242} is used to segment and track the visible arms in both ground truth (first row) and the generated frames (second row). The extracted segmentation masks (highlighted in \textbf{red}) are then used to assess the quality of body pose control with mIoU and Acc\%.
  }
  \label{fig:sam-viz}
\end{figure*}

\section{Experiments}
We use Nymeria~\cite{ma2024nymeria} for our experiments, which is the largest existing dataset of human motion captured `in the wild'. It features diverse individuals performing everyday activities across a wide range of real-world environments. Nymeria contains over 1,100 high-resolution egocentric video sequences ($1408\times1408$ at 30~FPS), each temporally synchronized with 3D body poses obtained from a full-body XSens motion capture system~\cite{xsens_ref}. 
We resample the videos to 16 frames per second and resize them to $480\times480$ resolution. From the full dataset, we selected a subset of 186 videos (approximately 50 hours in total) containing multiple participants and complex interactions. 
These videos were further divided into non-overlapping clips of 45 frames.

\subsection{Evaluation Protocol}
A comprehensive evaluation protocol is essential to assess whether an egocentric, pose-controllable video generation model can function as a reliable simulator of an embodied agent. To this end, we design an evaluation framework along three complementary dimensions that jointly capture visual quality, motion accuracy, and body control.

\paragraph{Visual quality and fidelity.}
We evaluate the overall realism and perceptual quality of the generated videos using standard metrics for video generation. Specifically, we report Fréchet Video Distance (FVD)~\cite{unterthiner2019fvd} for sequence-level evaluation, and the frame-level metrics SSIM, LPIPS, and DreamSim~\cite{fu2023dreamsim} to quantify perceptual similarity to the ground truth. We scale SSIM, LPIPS, and DreamSim by a factor of $100$ for easier comparison. These measures also provide an implicit indication of temporal consistency and adherence to pose guidance.

\paragraph{Global motion control.}
From an egocentric perspective, most visual changes correspond to the agent’s motion through the environment, which is effectively represented by the camera’s movement. Following prior work on camera-controllable video generation~\cite{he2024cameractrl}, we evaluate global motion control by measuring the average translation error (\texttt{TransError}) and rotation error (\texttt{RotError}). We used Vipe~\cite{huang2025vipe} to extract camera poses from both the ground-truth and generated videos. The full sequences, comprising both conditioning and predicted frames, were provided to Vipe, ensuring consistent scale alignment between the two pose trajectories.

\paragraph{Agent body control.}
While the previous metrics implicitly capture alignment between the generated video and the conditioning poses, they do not explicitly quantify the accuracy of full-body motion reproduction. To address this, we focus on evaluating visible body parts, primarily the arms, within the agent’s field of view. We employ SAM2~\cite{ravi20242} for video segmentation and tracking, applied to a random subset of sequences where the arms are visible. Keypoints were manually annotated in the first frame, and SAM2 was used to propagate the corresponding segmentation masks throughout the sequence as shown in Fig.~\ref{fig:sam-viz}. We then compute the mean Intersection-over-Union (IoU) between the ground-truth and generated arm masks, as well as the percentage of frames correctly detecting arm presence as a binary measure of body-control consistency.

\subsection{Implementation Details} 
We used Cosmos \cite{agarwal2025cosmos} as latent conditional video diffusion model and baseline. Cosmos is pretrained at scale on large video corpora (including \(\approx 8\%\) first-person view videos), making it a suitable backbone for our conditional egocentric video generation. We ran all our experiments using the \texttt{cosmos-predict2} (2B, 480p, 16 FPS) pre-trained model. We trained the model on videos of 45 frames, using 13 past frames as conditioning and generate 32 future frames (2 seconds) controlled by 3D full-body poses as described in the previous section. This model operates in a spatio-temporal compressed latent space using the Cosmos-Tokenizer, reducing the input to 12 latent frames.  
More details are in the supp.\ material and available at \href{https://cvg-bonn.github.io/EgoControl/}{\texttt{cvg-bonn.github.io/EgoControl}}.

\begin{table*}[t]
\centering
\resizebox{\linewidth}{!}{%
\begin{tabular}{c|ccc|c|cc|cc}
\toprule
\textbf{Experiments} & \multicolumn{3}{c|}{\textbf{Frame-level fidelity}} & \multicolumn{1}{c|}{\textbf{Video quality}} & \multicolumn{2}{c|}{\textbf{Motion control}} & \multicolumn{2}{c}{\textbf{Body control}} \\ 
 & SSIM$\uparrow$ & LPIPS$\downarrow$ & DreamSim$\downarrow$ & FVD$\downarrow$ & \texttt{TransError} & \texttt{RotError} & mIoU & Acc\% \\ 
\midrule
Base Cosmos & 42.29 & 50.62 & 23.00 & 71.00 & 16.53 & 15.60 & 20.03 & 85.36 \\
Finetuned & 47.47 & 45.74 & 18.14 & 40.70 & 9.93 & 13.65 & 25.13 & 85.20 \\
Head control & 56.94 & 29.71 & 10.22 & 22.68  & 5.16 & 3.29 & 33.70 & 91.14 \\
Body control & \textbf{58.60} & \textbf{26.71} & \textbf{8.54} & \textbf{20.18}  & \textbf{4.90} & \textbf{2.96} & \textbf{52.13} & \textbf{96.33} \\
\bottomrule
\end{tabular}
}
\caption{\model{} (row 4) shows notable improvements across all dimensions over the baseline and the head only control.}
\label{tab:main_results}
\end{table*}

\subsection{Results}
\paragraph{Quantitative results.}
\label{sec:quantitative_results}
\cref{tab:main_results} summarizes our main quantitative findings and performances of \model{}.
We report results for the pre-trained Cosmos backbone (row 1), then we establish a baseline by fine-tuning Cosmos using only past frames for context, with no pose guidance (row 2). 
Starting from this video prediction baseline, we introduce two variants: one conditioned solely on head motion ($\mathbf{\Delta h}$) and our final model (rows 3-4), which incorporates the complete full-body pose representation ($\mathbf{P}$) as defined in \cref{sec:pose_representation}.
Results show that, across all standard image- and video-level metrics, fine-tuning the pretrained backbone on Nymeria yields substantial gains over the off-the-shelf model. Using our proposed approach conditioned solely on head motion (row 3) substantially improves all metrics including body control further. The best results, however, are achieved when we control the video generation by the 3D full-body pose (row 4). The mIoU metric, which measures the alignment of the generated arms with the controlling arm poses, increases by nearly {55\%} relative to the head-only control. In over $96\%$ of the cases, the visibility of the arms in the generated frames is consistent with the ground-truth. It is very interesting to note that the full-body pose also improves the consistency of the camera view with the head pose compared to the head-only control. These results demonstrate that full-body information is critical to have full control over egocentric videos. Furthermore, the visual quality metrics (SSIM, LPIPS, DreamSim, FVD) are also improved by our model using full-body control.   

In \cref{tab:PEVA_comparison}, we provide a comparison with the results reported by PEVA~\cite{bai2025wholebody}.
Because PEVA generates instead of a video a single frame 2\,s in the future at $224\times224$ resolution given 15 past frames, we take the 32nd generated frame of \model{} and downsample it to match this horizon and image resolution. 
Under this protocol, \model{} outperforms PEVA despite using a shorter context window.
In \cref{tab:cami2v}, we also compare \model{} with full-body control to CAMI2V~\cite{zheng2024cami2v}, a camera control model finetuned on Nymeria.

\begin{table}[t]
\centering
\begin{tabular}{lccc}
\toprule
Model & LPIPS $\downarrow$ & DreamSim $\downarrow$ & FID $\downarrow$ \\
\midrule
PEVA XXL~\cite{bai2025wholebody} & 29.8 & 18.6 & 61.10 \\
\model{} & \textbf{24.3} & \textbf{11.3} & \textbf{50.68} \\
\bottomrule
\end{tabular}
\caption{Comparison with PEVA XXL. Note that PEVA uses 15 past frames at 4 FPS ($224\times224$) as conditioning and generates a single frame in the future. The comparison is for a generated frame 2 seconds in the future.}
\label{tab:PEVA_comparison}
\end{table}

\begin{table}[t]
\centering
\resizebox{0.8\linewidth}{!}{%
\begin{tabular}{c|c|cc|c}
\toprule
\textbf{Experiments}  & SSIM & \texttt{TransError} & \texttt{RotError} & mIoU \\ 
\midrule
CAMI2V~\cite{zheng2024cami2v} & 43.28 & 4.60 & 3.79 & 24.72 \\
EgoControl & \textbf{58.87} & \textbf{2.22} & \textbf{1.33} & \textbf{56.85} \\
\bottomrule
\end{tabular}
}
\caption{\footnotesize Comparison with the camera control method CAMI2V. Since CAMI2V generates fewer frames, we evaluate only the first 16 frames.}
\label{tab:cami2v}
\end{table}

\begin{figure*}
    \centering
    \includegraphics[width=1\linewidth]{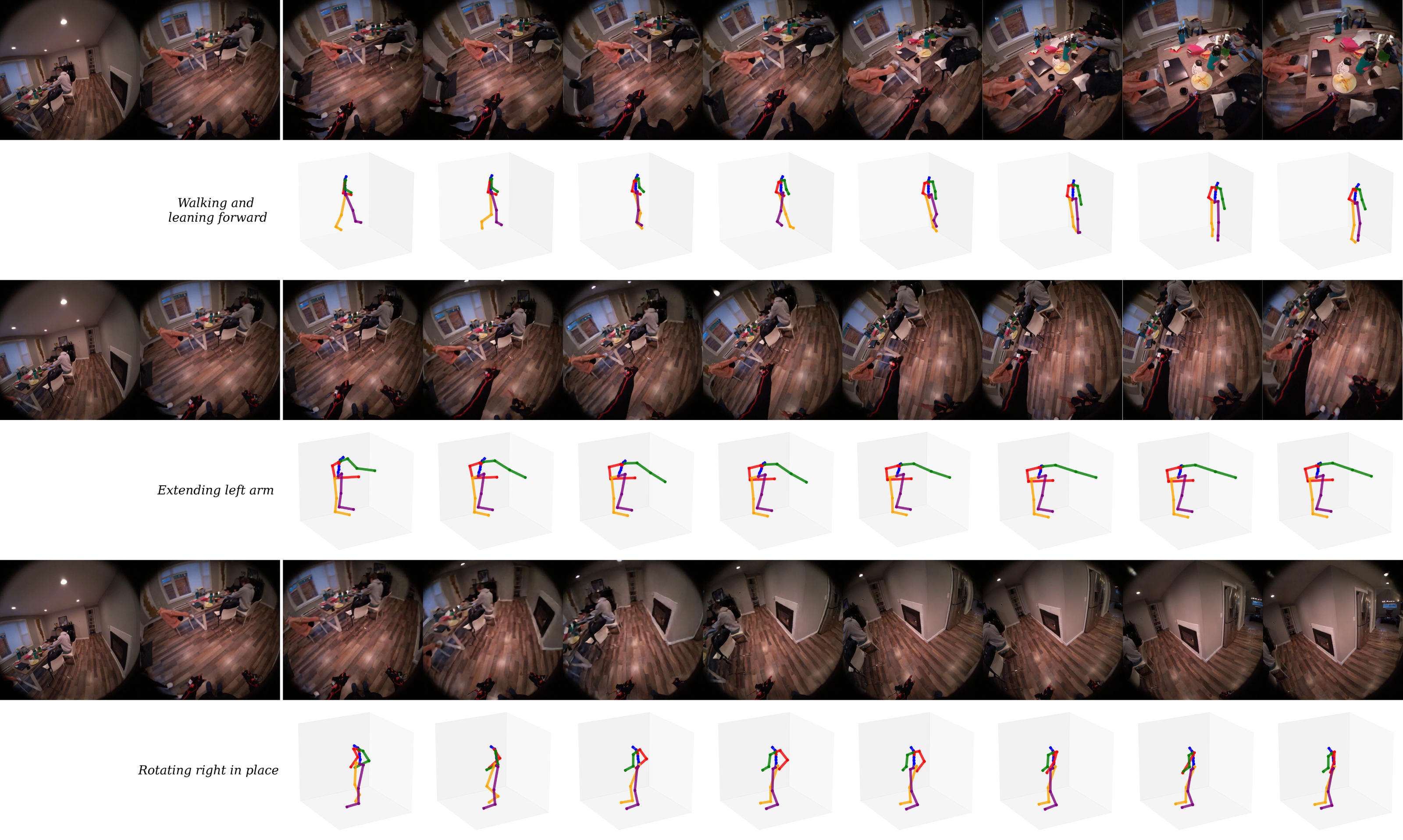}
    \caption{Given the same initial context, \model{} is capable of generating videos following different body movements. 
    See videos at \href{https://cvg-bonn.github.io/EgoControl/}{\texttt{cvg-bonn.github.io/EgoControl}} for better visualization.
    }
    \label{fig:same_context_diff_poses}
\end{figure*}

\begin{figure*}
    \centering
    \includegraphics[width=1\linewidth]{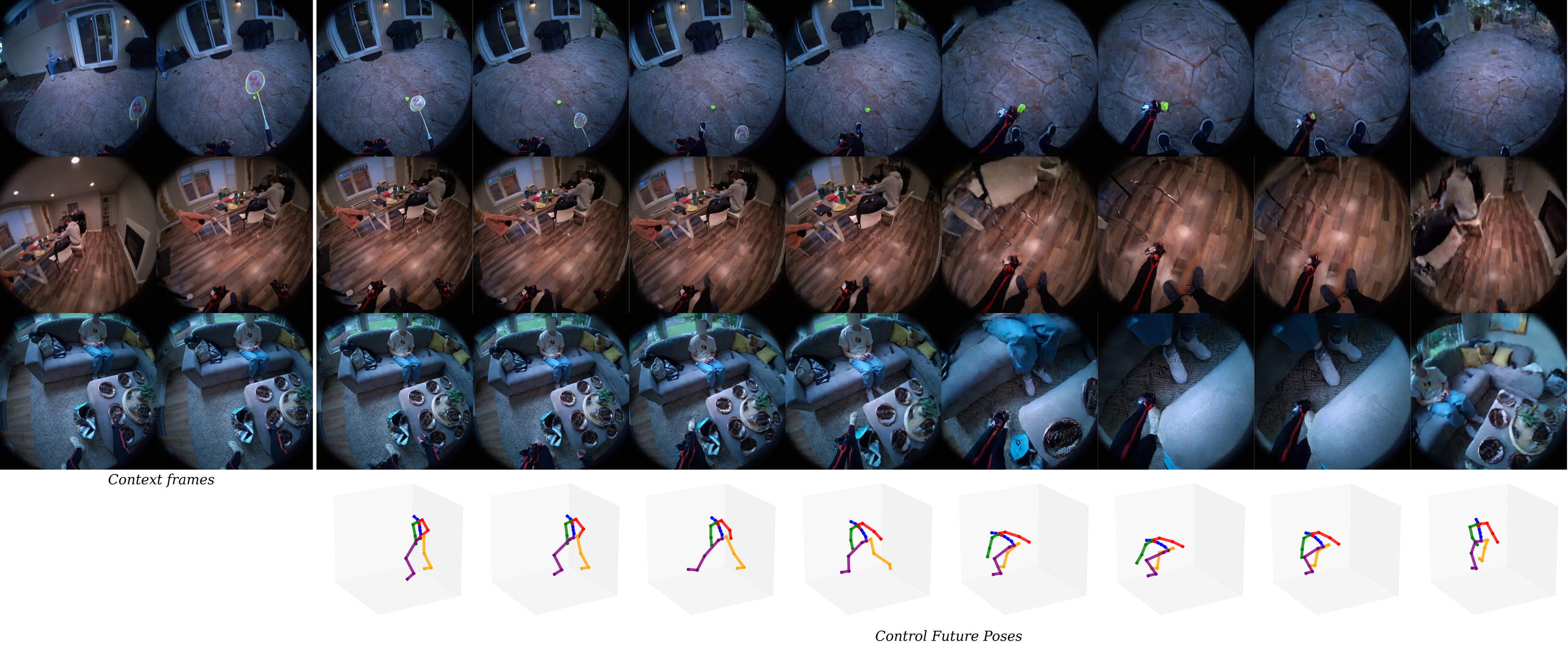}
    \caption{\model{} shows accurate pose alignment across three scenarios through different context frames. 
    See videos at \href{https://cvg-bonn.github.io/EgoControl/}{\texttt{cvg-bonn.github.io/EgoControl}} for better visualization.
    }
    \label{fig:same_poses_diff_contexts}
\end{figure*}

\paragraph{Qualitative results.}
Qualitatively, \model{} produces temporally coherent egocentric sequences that closely follow the controlling full-body pose. 
Fig.~\ref{fig:same_context_diff_poses} shows how different pose sequences, representing distinct actions like walking, rotating or moving an arm, can be applied to the same context frames, resulting in generated futures that follow the intended motions while adapting to the scene and objects.
Fig.~\ref{fig:same_poses_diff_contexts} further demonstrates the control abilities of \model{} by applying the same pose sequence to different initial contexts. In this case, the original pose of the video in the first row, representing the action of bending down to pick an object (a badminton ball), is applied to two different context frames in addition to the original one. In all three cases, the agent clearly reproduces the motion by bending down.

We provide more qualitative examples including object interactions in the supp.\ material and at \href{https://cvg-bonn.github.io/EgoControl/}{\texttt{cvg-bonn.github.io/EgoControl}}.
This also includes results at higher resolution ($960\times960$) and on Ego4D~\cite{grauman2022ego4d}.

\subsection{Ablation Studies}
\label{sec:ablation}
We conducted a series of ablation studies to validate our core design choices. We systematically investigated the efficacy of our proposed control mechanism and pose representation.

\begin{figure}[t]
    \centering
    
    \begin{subfigure}{0.8\linewidth}
        \centering
        \includegraphics[width=1\linewidth]{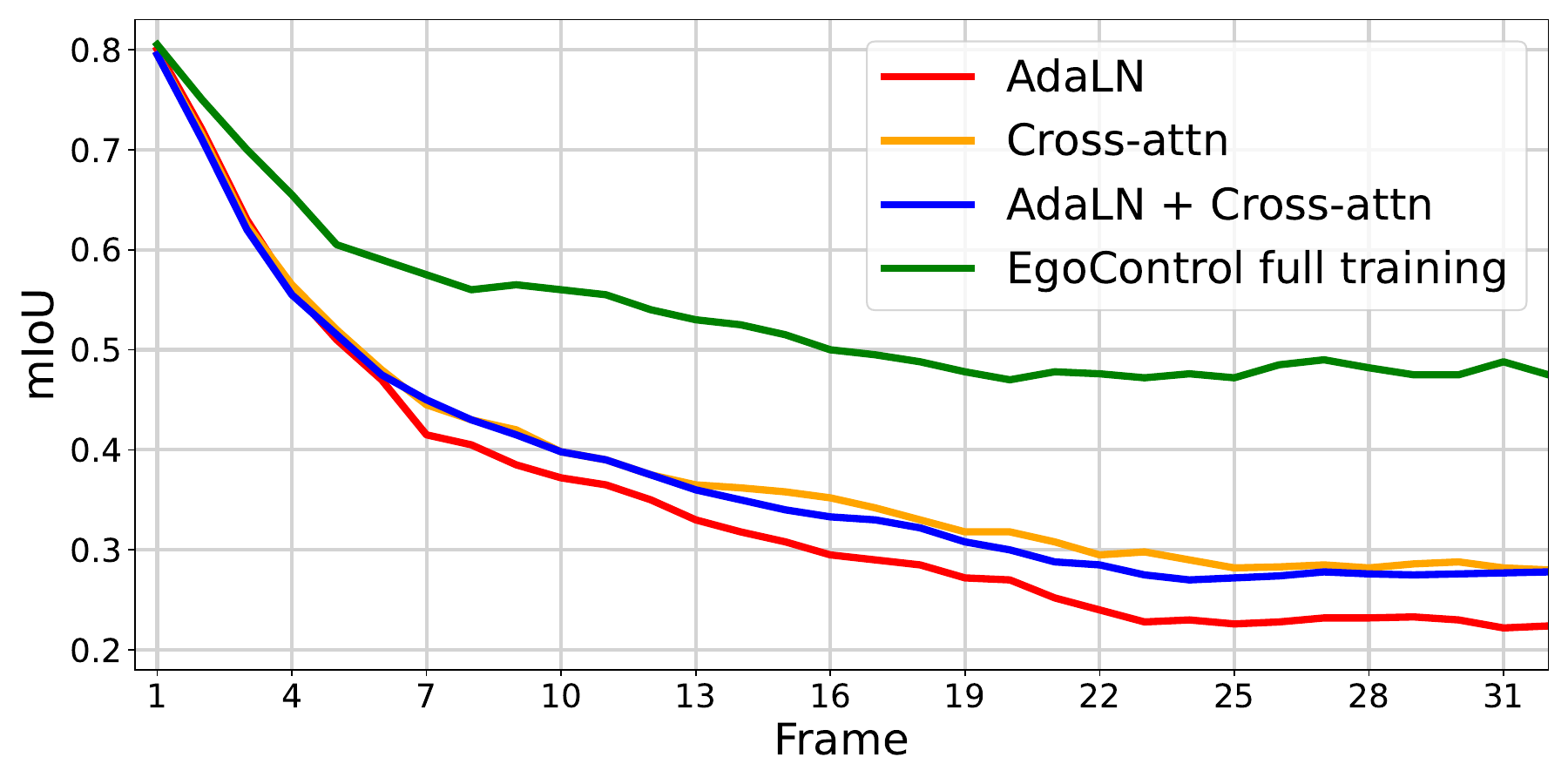}
        \caption{\footnotesize Average mIoU per frame.}
        \label{fig:miouframe}
    \end{subfigure}
    \begin{subfigure}{0.8\linewidth}
        \centering
        \includegraphics[width=1\linewidth]{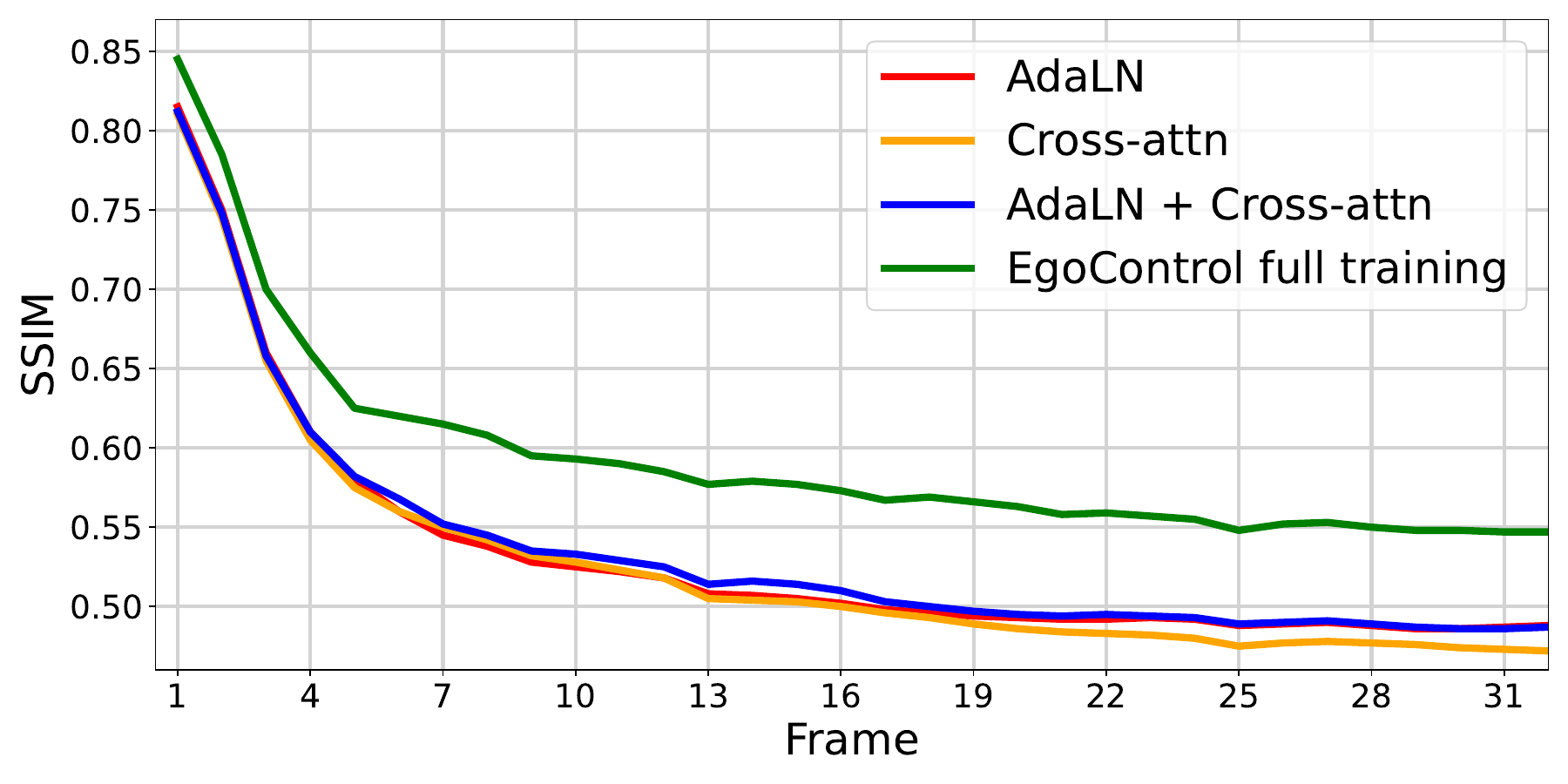}
        \caption{\footnotesize Average SSIM per frame.}
        \label{fig:ssimframe}
    \end{subfigure}

    \caption{Pose alignment (mIoU) and SSIM for each of the 32 generated frames. The green line denotes the fully trained model.}
    \label{fig:combined_plots}
\end{figure}

\paragraph{Control method.}
To justify our hybrid design, which integrates pose control via both modulation (AdaLN) and cross-attention, we evaluated three distinct variants in \cref{tab:control_ablation}. For the ablation studies, we trained the models with $7.000$ iterations, which is less than in \cref{tab:main_results}. When we compare conditioning with only AdaLN (row 1) and only cross-attention (row 2), we observe that AdaLN captures appearance and head motion better but struggles to enforce fine-grained articulated body control, i.e., it is better for the visual quality and global motion metrics (\texttt{TransError}, \texttt{RotError}) but it is worse in mIoU. The advantage of cross-attention for body alignment is expected since it utilizes the information of the controlling human pose sequence in a more granular way than AdaLN. The combination of both methods yields superior performance in both visual quality and motion accuracy, while being similar to using only cross-attention for body control (mIoU). Figs.~\ref{fig:miouframe} and \ref{fig:ssimframe} furthermore plot mIoU and SSIM for each generated frame separately, confirming the advantage of combining both control mechanisms. The plots also show the results for the fully trained model.

\begin{table}[t]
\centering
\resizebox{\linewidth}{!}{%
\begin{tabular}{lccc|c|c|cc}
\toprule
Control type & SSIM & LPIPS & DreamSim & FVD & mIoU & \texttt{TransError} & \texttt{RotError} \\
\midrule
AdaLN & \underline{52.16} & \underline{38.38} & \underline{11.48} & \underline{29.14} & 33.44 & \underline{6.07} & \underline{5.99}\\
Cross-attn (CA) & 51.65 & 38.89 & 12.20 & 29.28 & \textbf{37.84} & 6.85 & 7.19\\
AdaLN + CA & \textbf{52.60} & \textbf{36.79} & \textbf{10.94} & \textbf{27.51} & \underline{37.40} & \textbf{5.59} & \textbf{5.23}\\
\bottomrule
\end{tabular}
}
\caption{Ablation on the control mechanism.}
\label{tab:control_ablation}
\end{table}

\paragraph{Pose representation.}
The manner in which motion is represented is critical for effective conditioning. We explored several alternatives to our proposed formulation. 
First, for head-only control, we compared our differential representation (encoding relative transformations between consecutive frames) against a cumulative representation (encoding transformations relative to the initial pose $\mathbf{H}_0$). As shown in \cref{tab:pose_ablation}, the \texttt{TransError} of the cumulative representation is nearly twice as high compared to our differential strategy. While conceptually simpler, the cumulative approach suffers from numerical instability as motion values grow over long sequences. The differential formulation provides a more stable and temporally accurate signal, leading to superior performance in both visual (e.g. SSIM and FVD) and motion metrics.
Second, for full-body control, we validated our choice of representing joint locations relative to the pelvis. We compared this approach with an alternative that encodes the per-frame relative motion of each joint ($\mathbf{\Delta}\mathbf{j}$), similar to our representation of head motion. The results, also in \cref{tab:pose_ablation}, show that this frame-to-frame differential encoding for joints is suboptimal. The model struggles to maintain body coherence, as evidenced by a significant drop in pose control metrics. An improvement of {5.55} in mIoU confirms that a pelvis-centric coordinate system provides a more robust and easier-to-learn representation for articulated body dynamics.

\begin{table}[t]
\centering
\resizebox{\linewidth}{!}{%
\begin{tabular}{l|ccc|c|cc}
\toprule
\multicolumn{7}{c}{Head motion} \\
Head only variants & SSIM & LPIPS & DreamSim & FVD & \texttt{TransError} & \texttt{RotError} \\
\midrule
Cumulative & 54.51 & \textbf{32.69} & 11.08 & 27.23 & 12.31 & \textbf{4.11} \\
Differential & \textbf{55.16} & 32.99 & \textbf{10.86} & \textbf{24.01} & \textbf{6.75} & 4.67 \\
\midrule
\midrule
\multicolumn{7}{c}{Full-body pose representation} \\
Fullbody variants & SSIM & LPIPS & DreamSim & FVD & mIoU & Acc\%\\
\midrule
$\mathbf{P}=[\mathbf{\Delta}\mathbf{j]}\times23$ & \textbf{52.90} & \textbf{36.07} & 11.26 & 30.12 & 31.85 & 91.17\\
$\mathbf{P} = [\mathbf{\Delta} \mathbf{h}, \mathbf{\Delta} \mathbf{r}, \mathbf{J}]$ & 52.60 & 36.79 & \textbf{10.94} & \textbf{27.51} & \textbf{37.40} & \textbf{93.03}\\
\bottomrule
\end{tabular}
}
\caption{Ablation on the pose representation.}
\label{tab:pose_ablation}
\end{table}

\section{Conclusion}
We presented \model{}, a diffusion-based egocentric video generator that conditions future-frame synthesis on 3D full-body pose sequences of the camera wearer. By introducing a compact pose representation and a dedicated control pathway, \model{} translates global camera dynamics and articulated body motion into temporally coherent, pose-consistent egocentric videos, improving substantially both visual fidelity and motion alignment compared to baselines. At the same time, our study reveals clear limitations that point to promising future work. First, Nymeria does not provide explicit hand-pose annotations, so our current model does not explicitly model fine-grained hand articulation. While full-body conditioning substantially improves arm and coarse-hand motion, this gap limits accurate hand control.
Second, our training data is biased toward a specific acquisition setup with fisheye-mounted cameras and subjects wearing a motion-capture suit, which may reduce out-of-distribution generalization to everyday viewpoints, clothing, and sensor rigs. 
Future work should therefore extend pose conditioning to include explicit hand representations and utilize additional training data to broaden camera and clothing variability. 
We believe these directions will further close the gap between controllable egocentric synthesis and realistic embodied simulation for downstream planning and interaction tasks.

\section*{Acknowledgements}
This work was supported by the ERC Consolidator Grant FORHUE (101044724). We gratefully acknowledge the Federal Ministry of Research, Technology and Space, the Ministry of Culture and Science of the State of North Rhine-Westphalia, the Ministry of Science, Research and Arts of the State of Baden-Württemberg, the Bavarian State Ministry of Science and the Arts and the Gauss Centre for Supercomputing e.V. (GCS) for funding this project by providing computing time on the Supercomputer JUPITER at Jülich Supercomputing Centre (JSC) of Forschungszentrum Jülich through the Gauss AI Compute Competition. We also thank the German AI Service Center WestAI (01IS22094A) for providing additional computational resources.

{
    \small
    \bibliographystyle{ieeenat_fullname}
    \bibliography{main}
}

\clearpage
\setcounter{page}{1}
\maketitlesupplementary

\section{Additional Implementation Details}

We applied classifier-free guidance using the input context frames for the Cosmos baseline and \model{}. 
During training, the observed context frames were randomly dropped with a probability of 0.2, and we pre-extracted the latent embeddings to reduce computational overhead.  We used a global batch size of 1024, learning rate of $1.1\cdot10^{-5}$, and trained our final version of \model{} for $30.000$ iterations on H100s GPUs. At inference time, we enabled video guidance with a guidance weight of 2, which resulted in better qualitative results. We conducted all experiments without applying the default Cosmos Guardrail to the conditioning frames or the generated videos, ensuring that it did not influence the results. For the body control alignment evaluation, we used the \texttt{small} version of SAM2. 
The trained model and evaluation scripts are available at \href{https://cvg-bonn.github.io/EgoControl/}{\texttt{cvg-bonn.github.io/EgoControl}}. 

\section{Global Motion Evaluation}
While in the previously described experiments the global motion control evaluation is performed using VIPE to extract the camera motion from videos, we report in \cref{tab:vggt_results} results using VGGT~\cite{wang2025vggt} for extracting camera motion.
This evaluation shows similar relative improvements across our experiments, further showing the positive effects of full-body conditioning for effective control.

\section{Additional Qualitative Results}
\begin{figure*}
    \centering
    \includegraphics[width=1\linewidth]{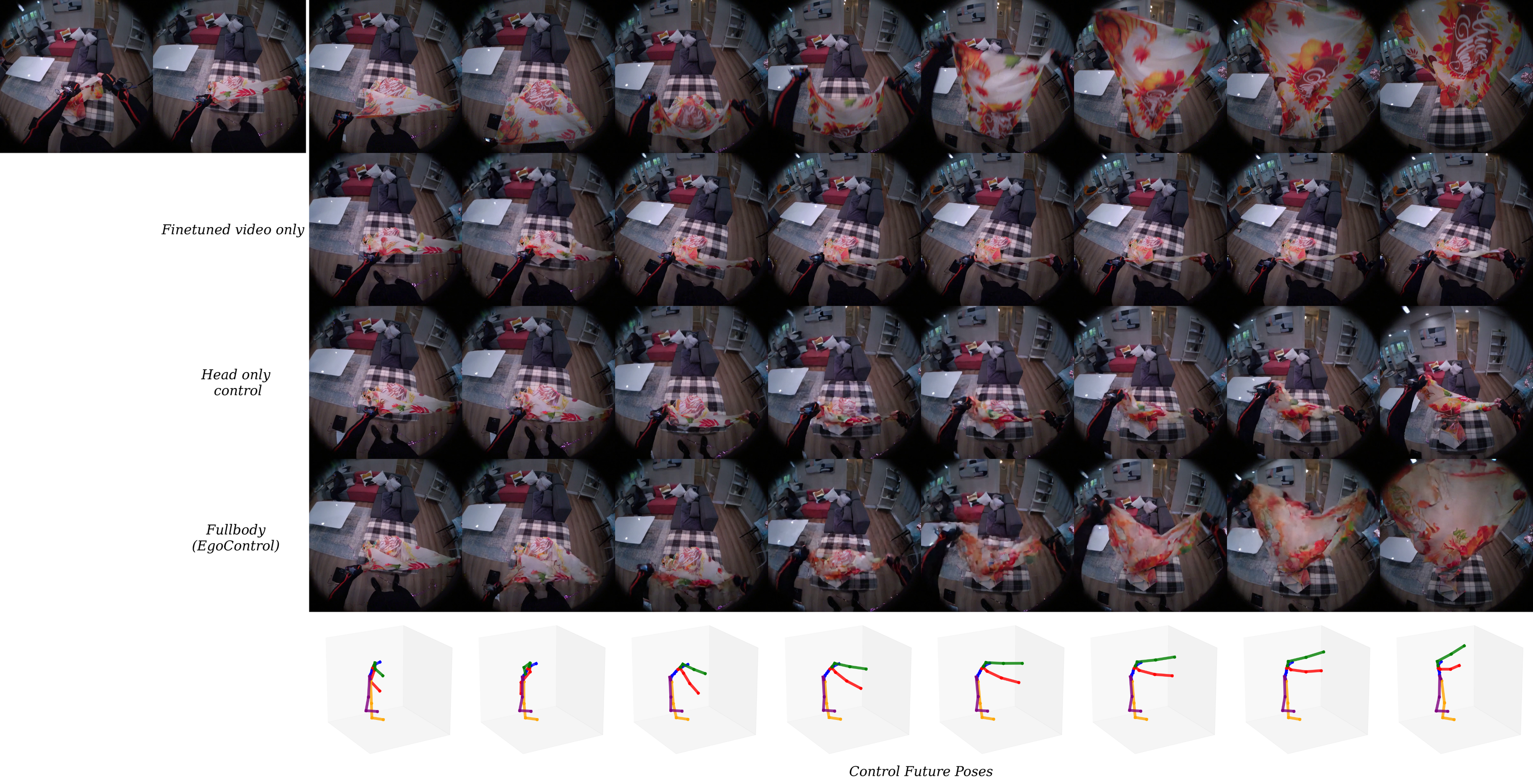}
    \caption{Comparing \model{} (fourth row) to ground truth (first row), finetuning (second row), and head only control (third row).}
    \label{fig:baseline_comparison}
\end{figure*}

We proceed describing some additional qualitative results.   
\paragraph{Baseline comparison.}
In order to show the effect of our full-body pose control in the generated videos, we include a visual comparison of models using different control information in Fig.~\ref{fig:baseline_comparison}. More specifically, it can be seen that the simple finetuned version of Cosmos (row 2), i.e., with no other information than the past context frames, does not follow the body pose and camera view compared to the ground truth video. This behaviour is expected due to multiple possible feasible futures given only the context frames. 
In row 3, we show the generated video using only the head pose for control.
The camera view starts to follow a similar path to the ground truth, however, the hand movements are still different from what they should be. 
Finally, by including the full-body pose (row 4), \model{} manages to control both the camera view and body movements, precisely generating the motion of the arms going upward and the complex interaction with the sheet. 

\begin{table}[t]
    \centering
    \resizebox{\linewidth}{!}{%
    \begin{tabular}{c|cc|cc}
         Experiment & \multicolumn{2}{c|}{VIPE} & \multicolumn{2}{c}{VGGT}\\
         & \texttt{TransErr} & \texttt{RotErr} & \texttt{TransErr} & \texttt{RotErr} \\
         \midrule
         Base Cosmos & 16.53 & 15.60 & 6.89 & 12.29   \\
         Finetuned & 9.93 & 13.65 & 5.01& 11.28\\
         Head Control & \underline{5.16} & \underline{3.29} & \underline{2.83} & \underline{3.46}\\
         Body Control & \textbf{4.90} & \textbf{2.96} & \textbf{2.40} & \textbf{2.76}\\
    \end{tabular}
    }
    \caption{Motion evaluation with VGGT instead of VIPE.}
    \label{tab:vggt_results}
\end{table}

\begin{figure*}
    \centering
    \includegraphics[width=1\linewidth]{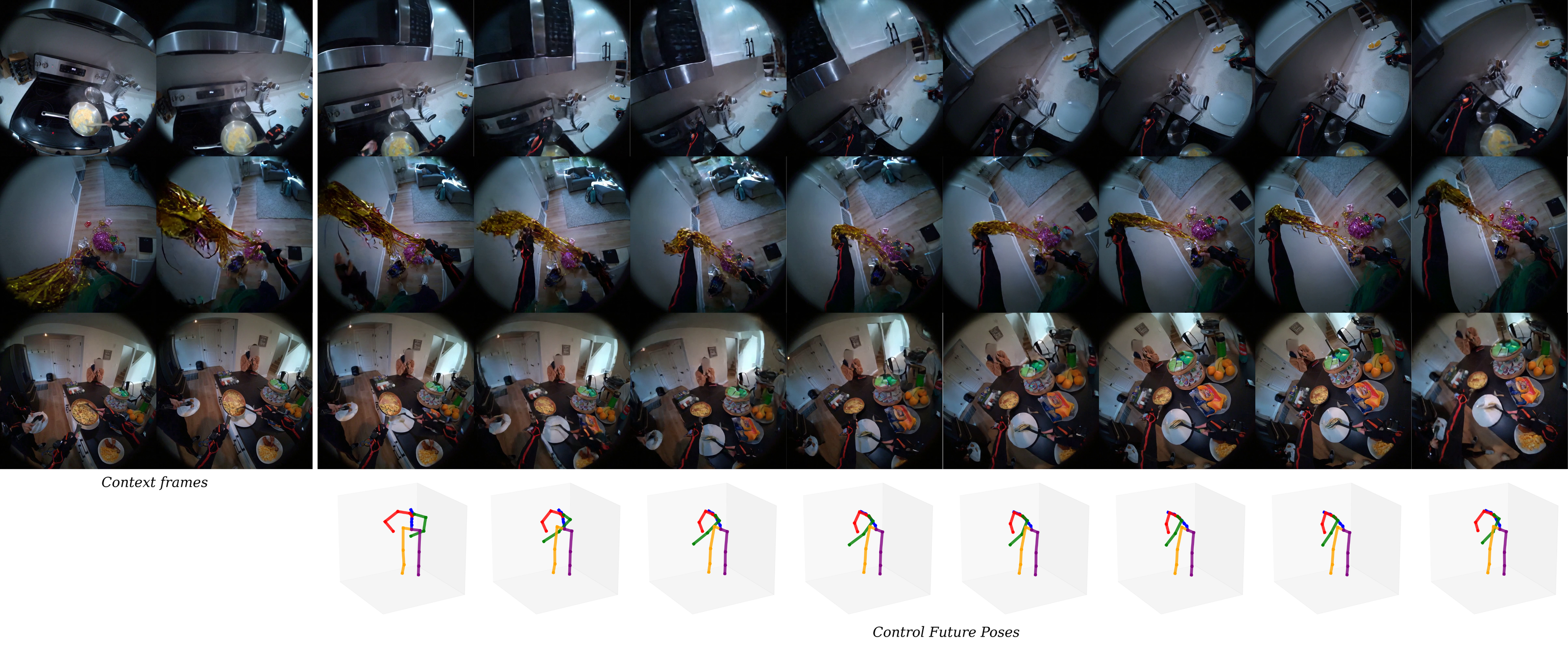}
    \caption{Applying the same sequence of human poses to different context frames. \model{} shows accurate pose alignment for all three scenarios.}
    \label{fig:same_pose_supp}
\end{figure*}

\paragraph{Different context.}
In Fig.~\ref{fig:same_pose_supp}, we show another example of applying the same sequence of poses, in this case extending the left arm while making a slight upper-body rotation, to three different initial context frames.
The results show that regardless of the initial state of the person, the generated video is successfully controlled by the conditioning pose. It should be noted that the location of the hands in the 2D pixel space is not necessarily the same across different contexts. Depending on the initial head pose, the view of the body can be different, given the correctly followed pose. 

\begin{figure*}
    \centering
    \includegraphics[width=1\linewidth]{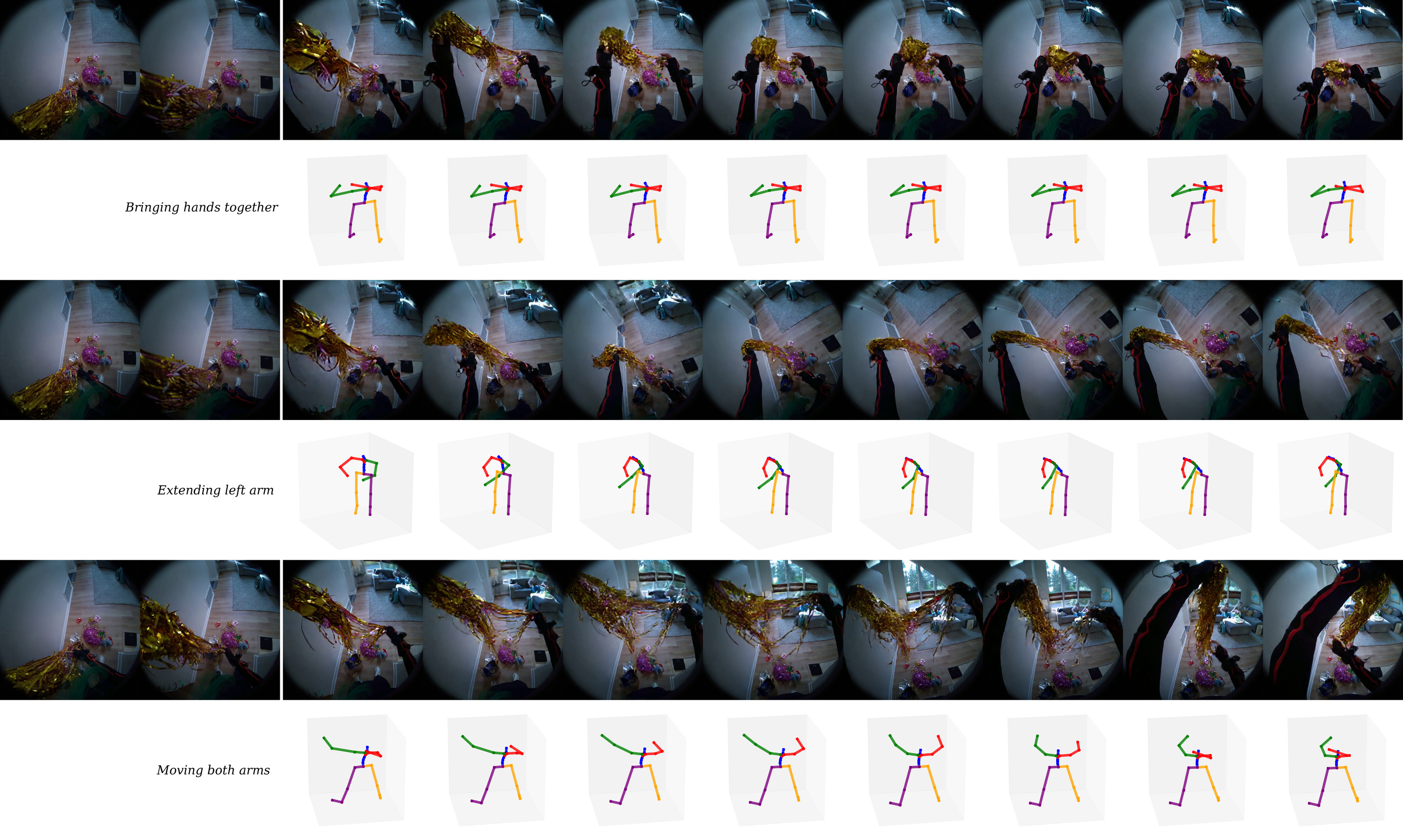}
    \caption{Applying different sequences of human poses to the same context frames. \model{} is capable of generating videos following the different body movements.}
    \label{fig:same_context_supp}
\end{figure*}

\paragraph{Different control pose.}
The last example of control abilities of \model{} is shown in Fig.~\ref{fig:same_context_supp} by applying different control poses to the same initial context frames.
In the first row, we apply a pose that differs significantly from the starting one in the context frames.
In this case, the model transitions smoothly toward the target pose, effectively bringing the hands together. 
The second and third row present other types of movements, including controlling only the left arm or both arms simultaneously.
Interestingly, these examples show a physically plausible interaction with the object being held.
\paragraph{Higher resolution.}
We finetune the final version of \model{} on videos at $960\times960$ resolution for additional $2.000$ iterations. This short finetuning stage enables \model{} to generate higher resolution videos. Examples are shown at \href{https://cvg-bonn.github.io/EgoControl/}{\texttt{cvg-bonn.github.io/EgoControl}}. 

\begin{figure*}
    \centering
    \includegraphics[width=1\linewidth]{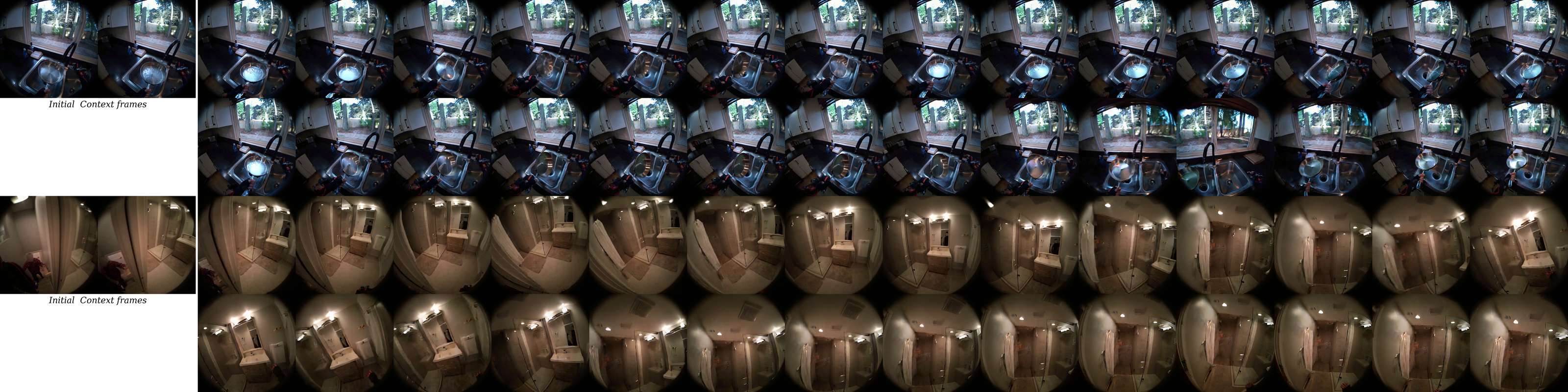}
    \caption{Videos of 8 seconds generated by \model{}.}
    \label{fig:long_video}
\end{figure*}

\paragraph{Longer video generation.}
To generate longer sequences, we run \model{} autoregressively by iteratively feeding the last generated frames back as conditions. Fig.~\ref{fig:long_video} shows two examples, each with a duration of 8 seconds.

\begin{figure*}
    \centering
    \includegraphics[width=1\linewidth]{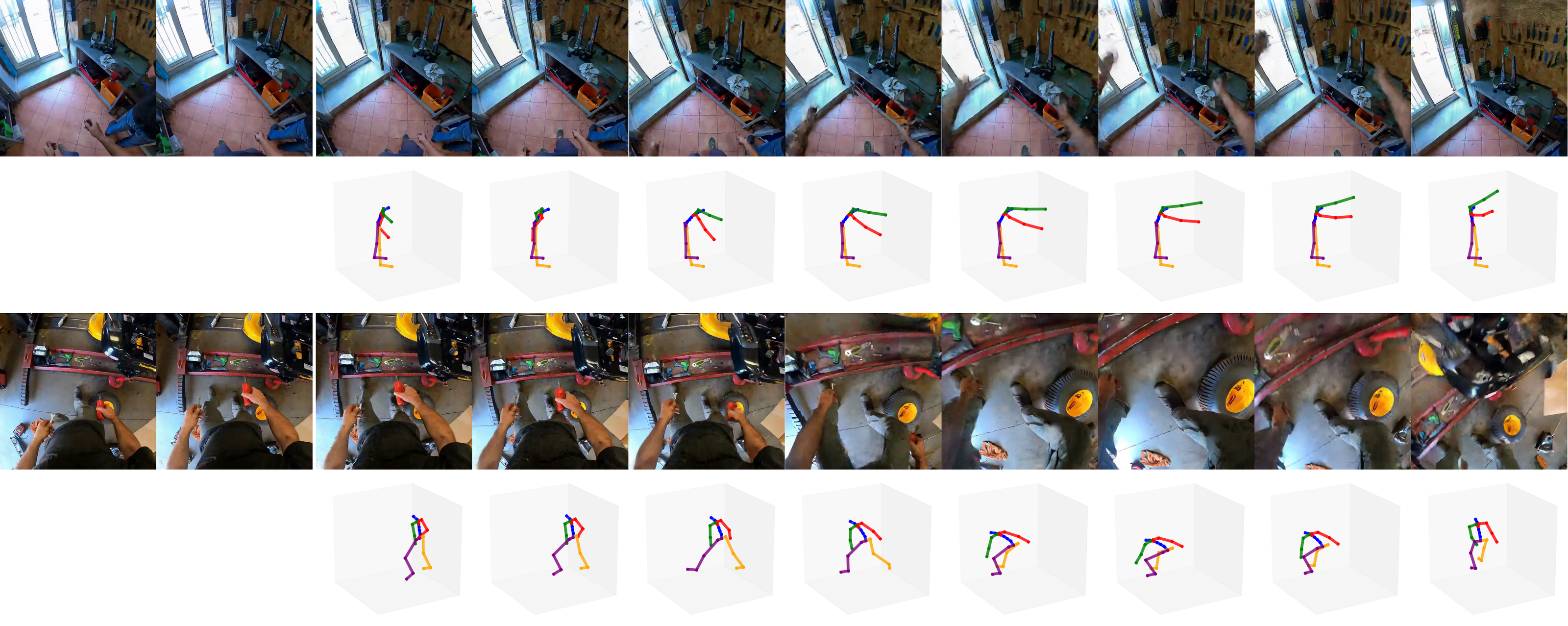}
    \caption{\model{} shows control abilities on Ego4D, a dataset without any pose annotations.}
    \label{fig:ego4d_qual}
\end{figure*}

\paragraph{Ego4D.}
To assess generalization, we also take a subset of 250 videos (200/50 train/val) of Ego4D~\cite{grauman2022ego4d}. The dataset, however, does not contain any body pose annotation. 
The final version of \model{} is thus finetuned on a subset of Ego4D without body poses. With just 1.000 iterations, the model generalizes, showing control abilities even with a new camera setting and body appearance (\cref{fig:ego4d_qual}).  

\begin{figure*}
    \centering
    \includegraphics[width=1\linewidth]{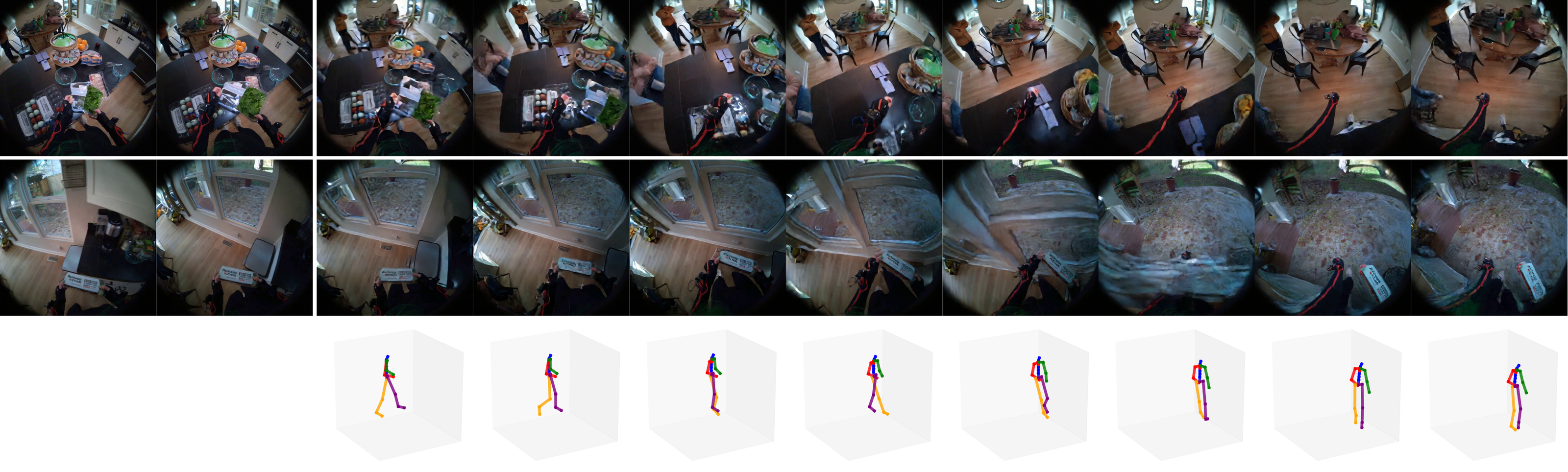}
    \caption{Failure cases when the provided sequence of poses is inconsistent with the environment.}
    \label{fig:failures}
\end{figure*}

\paragraph{Failure Cases.} 
\cref{fig:failures} shows some failure cases caused by physically implausible sequences of poses. In the first row, the person walks through the table. In the second row, the person carries an object and walks through a window. In most cases, implausible control sequences can be recognized by strong artifacts. For instance when the person walks through the window, the object and arm disappear. Automatically detecting physically implausible sequences of poses is an interesting future work.       

\end{document}